\documentclass[conference]{IEEEtran}
\IEEEoverridecommandlockouts
\usepackage{cite}
\usepackage{amsmath,amssymb,amsfonts}
\usepackage{algorithmic}
\usepackage{graphicx}
\usepackage{textcomp}
\usepackage{xcolor}
\usepackage{adjustbox}
\usepackage{graphicx}
\usepackage{subcaption}
\usepackage{gradient-text}
\usepackage{xcolor}
\usepackage{booktabs}
\usepackage{bm}
\usepackage{xspace}
\usepackage{siunitx}
\usepackage{float}

\newcommand{\our}{ADO\xspace}

\begin{document}

\title{\huge Balancing Accuracy and Efficiency: Adaptive Dynamics Orchestration for Model Predictive Control}

\author{\IEEEauthorblockN{Francesco Cancelliere\textsuperscript{1}, Aniket Datar\textsuperscript{2}, Giovanni Muscato\textsuperscript{1}, and Xuesu Xiao\textsuperscript{2}}
\IEEEauthorblockA{\textsuperscript{1}Department of Electrical, Electronic and Computer Engineering (DIEEI), University of Catania, Italy \\
\textsuperscript{2}Department of Computer Science, George Mason University, USA}}

\makeatletter
\g@addto@macro\@maketitle{
\begin{figure}[H]
  \setlength{\linewidth}{\textwidth}
  \setlength{\hsize}{\textwidth}
  \centering
    \includegraphics[width=\linewidth]{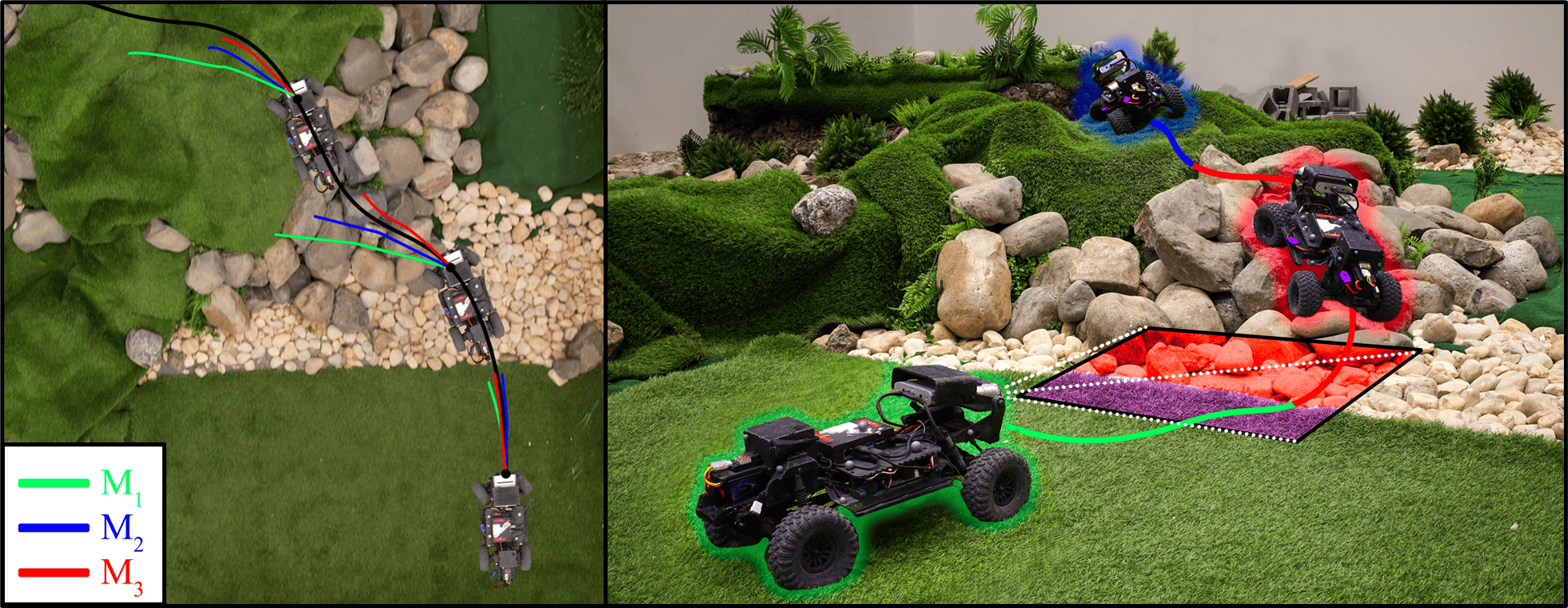}
    \caption{\textbf{Adaptive Dynamics Orchestration (\our)} adaptively selects among kinodynamic models ($M_1$--$M_3$) with different accuracy-efficiency trade-offs. 
    \textbf{Left:} accuracy comparison of model rollouts (green, blue, and red trajectories correspond to the least to most accurate, but most to least efficient models, respectively) against the executed trajectory (black). 
    \textbf{Right:} terrain-conditioned model switching (efficient models on easy terrain and accurate models on rough regions).}
    \label{fig::teaser}
 \end{figure}
}
\addtocounter{figure}{-2}
\maketitle
\begin{abstract}
Model Predictive Control (MPC) for autonomous navigation faces a fundamental trade-off between model accuracy and real-time efficiency. High-fidelity dynamics models can accurately predict complex vehicle–terrain interactions during trajectory rollouts, but incur significant computational cost, increasing inference latency and reducing control frequency. Conversely, lightweight models enable fast updates and dense sampling, yet may produce erroneous predictions under safety-critical conditions, potentially leading to catastrophic failures such as vehicle rollover.
To address this trade-off, we propose Adaptive Dynamics Orchestration (\our), a framework that dynamically selects the most appropriate dynamics model for the current navigation context. \our maintains a library of models spanning diverse accuracy–efficiency profiles and continuously refines terrain-conditioned performance estimates using residual errors from online counterfactual rollouts, where executed control actions are replayed across the model library to assess predictive discrepancy. These estimates guide model selection in real time, balancing computational efficiency and predictive accuracy.
Real-world experiments on an off-road ground robot demonstrate that \our significantly reduces modeling error compared to a fixed low-latency baseline, while approaching the accuracy of the highest-fidelity model without incurring its computational cost, resulting in more reliable and effective navigation in challenging terrain.
\end{abstract}

\section{Introduction}

Model Predictive Control (MPC), particularly sampling-based methods such as Model Predictive Path Integral (MPPI) control~\cite{williams2017model}, has emerged as a powerful framework for autonomous navigation in complex environments. By rolling out thousands of candidate trajectories using a forward kinodynamic model in real time, these planners can navigate high-dimensional state spaces while satisfying nonlinear constraints. Their performance critically depends on high-frequency control updates, which enable rapid refinement of trajectory distributions and convergence toward optimal actions. Consequently, navigation performance is fundamentally governed by two tightly coupled factors: the fidelity of the dynamics model used during trajectory rollouts and the inference time incurred when querying that model.

Safe navigation in complex terrain demands high-fidelity kinodynamic models capable of capturing nuanced vehicle–terrain interactions and system dynamics. In unstructured environments such as loose sand or steep inclines, accurate prediction requires accounting for additional Degrees of Freedom (DoFs), including height, roll, and pitch, as well as higher-order effects such as momentum and terrain-dependent dynamics. Rich perceptual inputs, such as elevation maps and semantic terrain features, further influence these predictions. During trajectory evaluation, these factors directly affect the planner’s ability to anticipate stability limits and safety-critical behaviors—phenomena that simplified models often fail to capture.

However, high-fidelity models are computationally expensive, creating a fundamental trade-off between physical accuracy and real-time responsiveness. Under fixed onboard computation, increasing model fidelity lengthens model query time, thereby reducing MPC sample density or increasing control latency. In contrast, simplified first-order models enable massive parallelization and high-frequency updates, which are advantageous on predictable, flat terrain. In such scenarios, higher control rates can compensate for modeling inaccuracies, allowing faster traversal due to reduced computation overhead and quicker reaction times.

To reconcile this trade-off, we propose Adaptive Dynamics Orchestration (\our, Fig. \ref{fig::teaser}), a framework that maintains a library of dynamics models spanning diverse accuracy–efficiency profiles and proactively selects the most suitable model for the current navigation context. Central to \our is an online counterfactual learning mechanism that enables the robot to autonomously characterize model performance during deployment, particularly in previously unseen or out-of-distribution terrain. At runtime, executed control actions are replayed across the model library to compute residual errors relative to observed trajectories, yielding empirical estimates of predictive performance. These estimates update a predictive semantic orchestrator that maps environmental features to model reliability, enabling proactive model selection that balances accuracy and efficiency in real time.
We validate \our through real-world navigation experiments on an off-road ground robot. Compared to fixed-model baselines, \our reduces modeling error while avoiding the unnecessary computational cost of always deploying the highest-fidelity model. This adaptive balancing translates into improved overall navigation performance across diverse terrain conditions.

\section{Related Work}
We discuss related work in MPC, dynamics models for navigation, and model adaptation in complex environments. 

\subsection{Model Predictive Control}
MPC is a framework for closed-loop planning under dynamics constraints and fundamentally relies on a forward model to predict future outcomes~\cite{mayne2000mpc}. Sampling-based MPC methods, such as MPPI~\cite{williams2017model}, avoid local linearization by optimizing over stochastic rollouts, enabling strong performance in highly non-linear, high-speed settings~\cite{williams2016mppi}. When using MPC for autonomous navigation, performance is governed by the real-time compute budget onboard a mobile robot: sample count, horizon length, exploration variance, and the number of optimization iterations per control step jointly determine convergence quality and control latency. Recent MPC extensions incorporate perception and learned context to improve real-world navigation or exploration in unknown environments~\cite{zhai2025pamppi,xiao2023performermpc}, but they still assume repeatedly querying a single model during rollout.

\subsection{Dynamics Models for Navigation}
Robot kinodynamic models for navigation span from low-order kinematics (e.g., bicycle, Ackermann, and differential-drive) to high-fidelity learned 6-DoF kinodynamics for complex terrain. Higher-fidelity models improve prediction accuracy in safety-critical environments under aggressive motion~\cite{han2024aggressive} and enable platform-aware planning by conditioning on local geometry and proprioception~\cite{roth2025fdm}. Recent off-road work explicitly models $\mathbb{SE}(3)$ consequences and competence~\cite{pokhrel2024cahsor}, improves efficiency via terrain-attentive embedding to minimize computation~\cite{datar2024tal}, and enhances robustness to distribution shift with physics-informed uncertainty~\cite{cai2025pietra}. However, these gains come with increased per-rollout inference cost. Under fixed compute, expensive models reduce MPC samples and iterations or cause increased control latency, while overly simple models frequently require corrective replanning due to modeling error. 
Furthermore, it is difficult to obtain a single model that is accurate across all navigation scenarios. 

\subsection{Model Adaptation}
To tackle the inaccuracies of the model or out-of-distribution scenarios, model adaptation has been shown to be effective. While offline adaptation focuses more on adapting to different embodiments with limited data~\cite{hu2025carol}, online adaptation methods aim to improve model accuracy on-the-fly, often by using onboard sensing or learning to adapt parameters to new scenarios~\cite{xiao2021learning, karnan2022viikd,levy2025meta,lu2025decremental,lu2025adaptive}. Meanwhile, universal dynamics models seek broad generalization across vehicles and environments~\cite{xiao2025anycar}. While effective, these approaches typically retain a single model class and thus fixed, and oftentimes high, inference complexity to cover a variety of scenarios during deployment. 
However, the navigation benefit of model adaptation to achieve higher accuracy is terrain-dependent: the performance gains from complex, high-fidelity models often diminish as the environment becomes simpler, whereas the benefit of higher iteration speed and more and longer rollouts per planning cycle diminish as the terrain becomes more complex and model bias becomes the dominant failure mode. 

Within an MPC setup, \our treats the dynamics model as a decision variable and selects from a diverse library ranging from highly efficient to highly accurate models based on the current navigation scenario, instead of adapting in a fixed model class with fixed complexity. Therefore, \our explicitly trades accuracy against efficiency to optimize overall navigation performance under real-time, onboard constraints.

\section{Preliminaries}
We present preliminaries on MPC and the computational complexity of kinodynamics models in MPC, leading to the trade-off between accuracy and efficiency. 

\subsection{Model Predictive Control}

MPC for navigation computes a control policy by solving a finite-horizon optimal control problem at each control step. Given the current timestep $t$, a current vehicle state $\textbf{x}_t= [x_t,y_t,z_t,\phi_t, \theta_t, \psi_t, \Dot{{x}}_t, \dots] \in \mathcal{X}$, where $\mathcal{X}$ is the state space of 6-DoF vehicle configurations (translational $x$, $y$, $z$ and rotational roll $\phi$, pitch $\theta$, yaw $\psi$) along with necessary higher-order derivatives, and a local environment state $\textbf{m}_t$ (which encapsulates terrain elevation, semantic context, and other related information), the objective is to find an optimal control sequence $\bm{\tau}_t^* = [u_t, u_{t+1}, \dots, u_{t+H-1}] \in \mathcal{T}$, where $\mathcal{T}$ and $\mathcal{U}$ denote the sets of all possible trajectories and robot actions respectively, over a prediction horizon $H$. This sequence minimizes a cumulative cost function:
\begin{equation}
\begin{aligned}
\bm{\tau}^*_t = &~ \arg\min_{\tau_t}
\sum_{j=t}^{t+H-1} c(\textbf{x}_j, \textbf{u}_j) + c_\textrm{term}(\textbf{x}_{t+H}),\\
\text{s.t.}\;&\quad \textbf{x}_t = \hat{\textbf{x}}_t,\\
&\quad \textbf{x}_{j+1} = f(\textbf{x}_j, \textbf{u}_j, \textbf{m}_j), \quad j=t,\dots,t+H-1, \\
&\quad \textbf{x} \in \mathcal{X},\quad \textbf{u} \in \mathcal{U}, \\
\end{aligned}
\nonumber
\end{equation}
where $\hat{\textbf{x}}_{t}$  is the initial state, $c$ is the step cost function, $c_\textrm{term}$ is the cost of the terminal state, and $f$ is the forward kinodynamic model that maps $\textbf{x}_j$, $\textbf{u}_j$, $\textbf{m}_j$ to the next $\textbf{x}_{j+1}$.

In complex environments, the true forward dynamics $f$ is highly non-linear and non-convex. Consequently, solving this optimization analytically or via standard gradient-based solvers is computationally prohibitive for real-time deployment and highly susceptible to local minima. 

To circumvent this, sampling-based MPC approaches, such as MPPI control~\cite{williams2017model}, approximate a near-optimal solution. Instead of analytically deriving $\bm{\tau}^*$, these planners sample thousands of candidate control sequences in parallel. By rolling out the forward kinodynamic model $f$ many times for each sequence, the planner evaluates the cost across all trajectories and computes a cost-weighted average to determine the optimal immediate command. 

\subsection{Computational Complexity of Kinodynamic Models}
\label{sec::complexity}
Because the planner evaluates thousands of potential future states at every timestep, the navigation performance of the autonomous system is fundamentally bottlenecked by two factors: the physical accuracy of the forward model $f$ and the computational time required to query it.

While the MPC optimization objective remains constant, the computational cost of rolling out the forward model $f(\textbf{x}_t, \textbf{u}_t, \textbf{m}_t)$ varies drastically depending on the required physical fidelity. On predictable, flat surfaces, simplified models with first-order kinodynamics enable massive parallelization and high-frequency updates. However, in unstructured terrain like loose sand or steep inclines, MPC performance is increasingly dictated by higher DoFs, higher-order kinodynamic derivatives, and richer perceptual inputs. Incorporating these elements fundamentally alters the computational complexity of the rollout phase in three primary ways:
\begin{itemize}%
\item State Space Dimensionality: Expanding the state representation from a planar approximation to a full $\mathbb{SE}(3)$ manifold increases the size of the underlying system of equations. Evaluating additional Degrees of Freedom (DoFs), such as height, roll, and pitch, necessitates more matrix operations per rollout time step, increasing the processing time for every sampled trajectory.
\item Sequential Time Dependencies: Simple kinematic models often allow the entire prediction horizon to be evaluated in a single, batch-parallelized call. Conversely, models incorporating higher-order derivatives or learned residual dynamics introduce strict sequential dependencies. Because each state prediction relies on the velocity outputs of the immediately preceding step, the horizon must be rolled out sequentially. This neutralizes the primary advantage of parallel compute architectures, constituting a primary source of additional latency.
\item Environmental Perception Bottlenecks: Integrating the local environment state $\textbf{m}_t$ requires mapping proprioceptive states to spatial representations. Whether geometrically querying a 2.5D elevation map or processing a local terrain geometry embedding, the memory bandwidth and inference time required to evaluate $\textbf{m}_t$ at every sample step significantly inflates the control loop latency.
\end{itemize}

\begin{figure*}[ht] %
\centering
\includegraphics[width=\linewidth ]{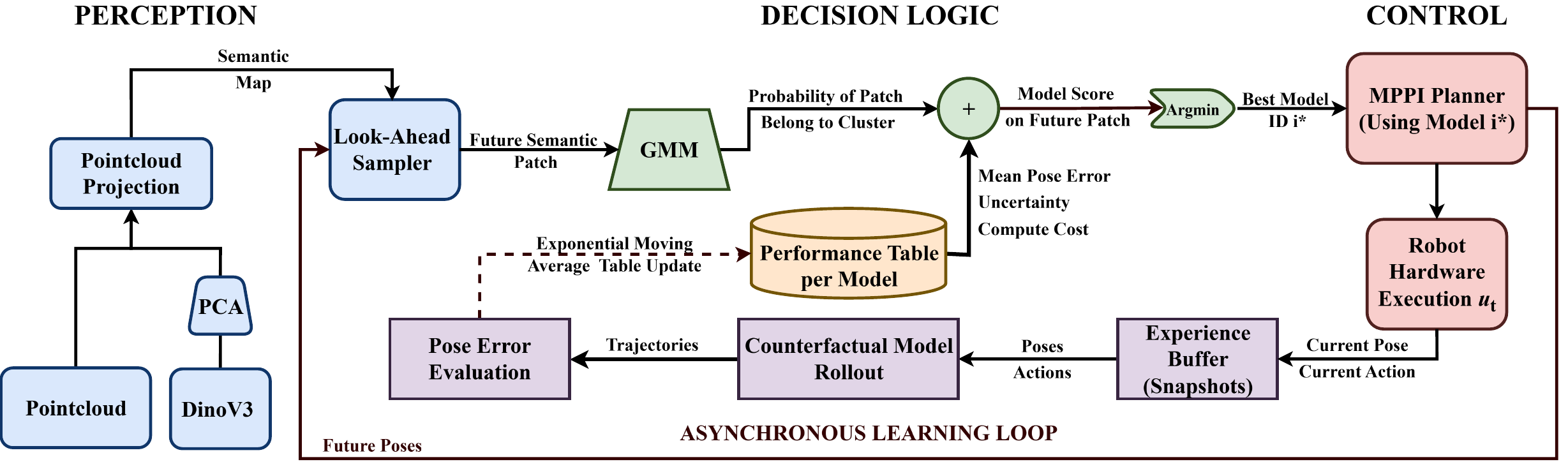}
\caption{Proactive Adaptive Dynamics Orchestration (ADO) framework featuring a visual-semantic perception pipeline (blue), an asynchronous counterfactual learning loop for model evaluation (purple), and a predictive gating module (green) that dynamically selects the most appropriate kinodynamic model for MPPI control (red).}
\label{fig:architecture}
\end{figure*}

\begin{figure}[ht] %
\centering
\includegraphics[width=\linewidth ]{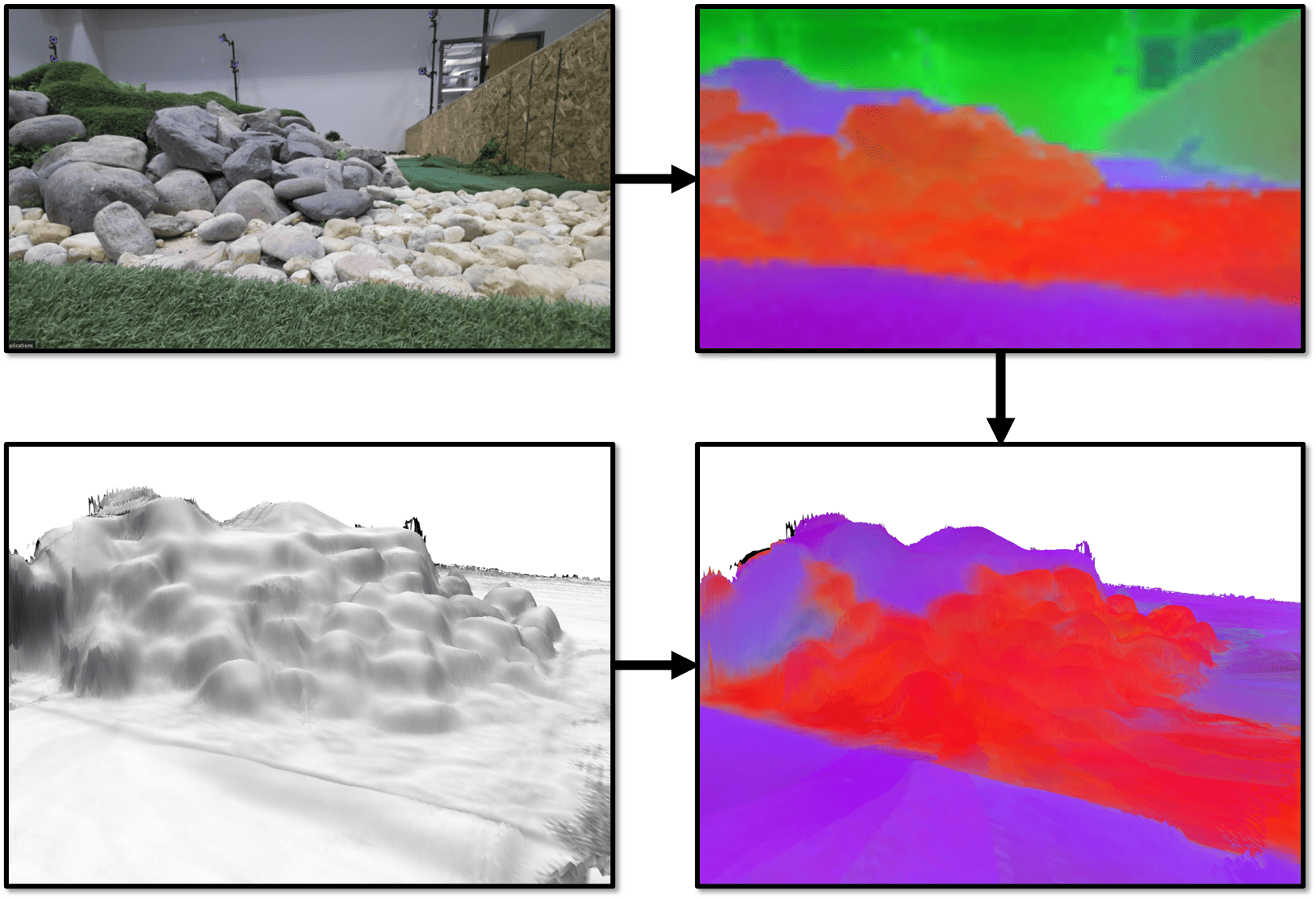}
\caption{Visual-semantic pipeline: the RGB image (top left) is converted into DinoV3-PCA features (top right), which is fused with the elevation map (bottom left) into the semantic elevation map (bottom right).}
\label{fig:dino_map}
\end{figure}

\section{Methodology}

Given a fixed computation budget, prioritizing model fidelity increases model query time, which inherently necessitates a reduction in MPC sample density or an increase in control latency. Because highly accurate models are unnecessarily slow for simple terrain, and fast, less accurate models inherently fail to capture kinodynamic consequences like stability and safety on complex terrain, a single model is typically insufficient to maximize navigation performance. 

To resolve this trade-off, our \our framework utilizes a library of kinodynamic models $\mathcal{M} = [M_1, M_2, M_3, \dots]$ spanning this Pareto front. By maintaining instantiations ranging from heavily parallelizable first-order kinematics to sequentially dependent, map-conditioned neural networks, the system can selectively deploy the appropriate physics formulation.

\subsection{Reactive Model Evaluation and Selection}

The \our framework centers on a counterfactual evaluation mechanism that continuously benchmarks the predictive competence of the models in the library. Instead of relying solely on the tracking performance of the active model, the system leverages the ability to replay previously executed control actions and their corresponding vehicle states across every model $M_i$. This allows the robot to autonomously characterize the accuracy of models that are not currently in control by simulating the trajectories they would have generated given the identical history of states and actions.

To leverage these counterfactual rollouts for performance estimation, the system measures the discrepancy between each model’s predicted trajectory and the ground truth recorded over a window of previous measurements. For a given model $M_i$, the orchestrator initiates a rollout from a historical state within the window and applies the recorded control sequence to produce a predicted trajectory. This trajectory is then compared against the actual path traversed by the robot to calculate a residual error $e_i$, which accounts for compounding open-loop divergence through a temporal discount factor:

\begin{equation}
e_{i} = \frac{1 - \lambda}{1 - \lambda^N} \sum_{j=1}^{N} \lambda^{j-1} \left( e^{\textrm{pos}}_{i, j} + e^{\textrm{rot}}_{i, j} \right),
\nonumber
\end{equation}
where $e^{\textrm{pos}}_{i, j}$ and $e^{\textrm{rot}}_{i, j}$ are the position and rotation error for the $i^\text{th}$ model at the $j^{\text{th}}$ timestep and $\lambda$ is a discount factor that exponentially down-weights errors at later timesteps in the horizon.

The orchestrator tracks at inference time the error $E_{i}$ and the variance $V_{i}$ per model using an Exponential Moving Average (EMA), such that:
\begin{align}
\label{eq:Ei}
{E}_{i} &\leftarrow (1-\eta){E}_{i} + (\eta) \cdot e_{i},\\
\label{eq:Vi}
V_{i} &\leftarrow (1-\eta) V_{i} + (\eta) \cdot (e_{i}-{E}_{i})^{2},
\end{align}
where $\eta$ is the learning rate. Similarly, it tracks the computational cost $C_i$ (average inference time). The historical mean error ${E}_{i}$ and variance $V_{i}$ are updated at the control loop rate for the active model, whilst these metrics are updated at a lower frequency for the inactive models. 
At every step, the system computes a generic selection score $J_i$ for each model:
\begin{equation}
J_{i} = \alpha \cdot {E}_{i} + \beta \cdot V_{i} + \gamma \cdot C_{i}, \nonumber
\end{equation}
where $\alpha$, $\beta$, and $\gamma$ are weighting coefficients that prioritize accuracy, uncertainty, and efficiency, respectively. 
The model minimizing $J_i$ is then selected for the next control iteration.
This strategy maintains low overall computational overhead while ensuring the system can switch instantly if the error of the current model grows rapidly.

\subsection{Proactive Semantic Orchestration}

While the reactive formulation dynamically ranks models, it suffers from a critical limitation: it only adjusts the model preference after the robot has already traversed new terrain and accumulated substantial tracking error. To ensure safe navigation, the system needs to anticipate future terrain changes and proactively switch models before the tracking error grows.

To anticipate environmental changes, we utilize a visual-semantic pipeline that treats environmental perception as a weighting mechanism for model selection. Front-facing camera inputs are processed to produce a compact feature vector ${\bm{\xi}}$ that captures the primary textural and geometric variances of the upcoming terrain. Rather than maintaining a single global error tracking variable ${E}_{i}$ across all environments, we employ a pre-trained probabilistic classifier to soft-cluster the terrain into $K$ distinct classes and tracks the error ${E}_{i,k}$ per class.

For a given visual feature $\bm{\xi}$, this module outputs a responsibility weight $P(k|\bm{\xi})$, representing the probability that the terrain belongs to cluster $k$. This responsibility vector acts as a weighting mechanism, transforming the reactive EMA into a context-aware Model Competence Map. 
Extending Eqns.~\eqref{eq:Ei} and ~\eqref{eq:Vi}, the historical mean error ${E}_{i,k}$ and variance $V_{i,k}$ for model $i$ in terrain cluster $k$ are updated seamlessly based on the visual context recorded at the beginning of the evaluation window:
\begin{align}
{E}_{i,k} &\leftarrow (1-\eta P(k|\bm{\xi})){E}_{i,k} + (\eta P(k|\bm{\xi})) \cdot e_{i}, \nonumber \\
V_{i,k} &\leftarrow (1-\eta P(k|\bm{\xi}))V_{i,k} + (\eta P(k|\bm{\xi})) \cdot (e_{i}-{E}_{i,k})^{2}. \nonumber
\end{align}

Finally, by spatially projecting the camera features onto a global Semantic Elevation Map (Fig.~\ref{fig:dino_map}), the robot can sample the upcoming visual context $\bm{\xi}_\textrm{future}$ along its candidate trajectories. The predictive selection score $J_{i}$ becomes a weighted sum over the anticipated terrain clusters:
\begin{equation}
J_{i} = \sum_{k=1}^{K} P(k|\bm{\xi}_\textrm{future}) \cdot [\alpha \cdot {E}_{i,k} + \beta \cdot V_{i,k} + \gamma \cdot C_{i}].
\label{eq:score}
\end{equation}
By evaluating this score, the orchestrator proactively selects the optimal model $i^{*} = \arg\min_{i} J_{i}$ based on the terrain the robot is about to encounter.
The whole architecture is represented in Fig.~\ref{fig:architecture}.

\section{Implementations}

The specific kinodynamic models in our library are deliberately designed to exhibit a distinct, relative spread across the accuracy-efficiency Pareto front. This relative variance is what allows \our to demonstrate meaningful selection behavior, demonstrating the efficacy of the framework.

\subsection{Navigation System}
\subsubsection{Vision and Semantic Pipeline} 
The Predictive Semantic Orchestration relies on DINOv3-VIT-S to extract dense visual features from the Azure Kinect RGB stream. To minimize latency, these embeddings are projected into a 3-dimensional latent space using a pre-computed PCA matrix, which captures the majority of the variance in our test environment. The probabilistic terrain classification is performed by a Gaussian Mixture Model (GMM), calibrated offline using a representative dataset of the environment's visual textures.
While we utilize DINOv3 and a GMM for this implementation, \our is agnostic to the specific choice of encoder and classifier, requiring only a probability distribution over terrain contexts.

\subsubsection{MPPI Configuration and Cost Formulation} The MPPI planner operates at a control frequency of \SI{50}{\hertz}. At each step, the planner samples 2000 trajectories over a prediction horizon of $H = 30$ steps ($4.5$ seconds). The running cost $c(\textbf{x}_j, \textbf{u}_j)$ penalizes deviation from a globally planned path and excessive roll/pitch angles to prevent rollover, while the terminal cost $c_\textrm{term}(\textbf{x}_{t+H})$ evaluates the final predicted state at the end of the horizon, encouraging trajectories that make progress toward the goal and terminate in a favorable state.

\subsubsection{Mapping Formulation} The 2.5D elevation map (Fig. \ref{fig:dino_map}) is maintained dynamically to build a robot-centric elevation map, along with projecting $\bm{\xi}$, providing the information needed for the Predictive Semantic Orchestration. 

\subsection{Models}

\subsubsection{$\mathbb{SE}(3)$ Bicycle Model ($M_1$)}

The first model ($M_1$) is the most efficient in the library, relying on first-order kinematics and single-pass elevation map lookups to enable high sample counts and rapid replanning. In order to avoid the State Space Dimensionality, the model decomposes the full $\mathbb{SE}(3)$ state by estimating $x$, $y$, and $\psi$ via planar bicycle kinematics, based on random throttle and steering inputs, whilst  $z, \phi, \text{ and }\theta$ via plane fitting at the four predicted wheel-terrain contact points, similar to ~\cite{datar2025learning}. Since this model does not rely on previous information, it is highly parallelizable across the horizon, reducing inference time significantly.
While highly effective for flat or uneven terrain, $M_1$ does not account for complex dynamic interactions or high-speed maneuvers where first-order approximations may fail. 

\subsubsection{Bicycle + Learned Residual Model ($M_2$)}

The $M_2$ model functions as a hybrid system that augments traditional physics with machine learning. It starts with a first-order bicycle kinematic prior to handle fundamental robot movement. To improve accuracy, it incorporates a learned Multi-Layer Perceptron (MLP) specifically designed to predict the residual to the kinematic model. 
The MLP only needs to predict the \emph{correction} to the kinematic model due to higher-order kinodynamic derivatives, rather than the full dynamics.

Unlike $M_1$ which is parallelizable, $M_2$ suffers from Sequential Time Dependencies: each step's predicted residual depends on the previous step's output. This sequential dependency constitutes the primary source of additional latency compared to $M_1$, but enables $M_2$ to capture velocity-dependent effects, such as transient steering response that the pure kinematic model $M_1$ misses.

\subsubsection{Map-Conditioned End-to-End Model ($M_3$)}

The third model ($M_3$) incorporates an elevation-map-conditioned neural network that processes terrain data through a convolutional autoencoder to produce a compact embedding~\cite{datar2024tal}. This learned representation, combined with the current robot state and control action, allows the network to predict the subsequent state in  $\mathbb{SE}(3)$.

Because the $M_3$ network predicts the next state based on terrain elevation, the model can capture terrain-dependent dynamics, such as slip or speed reduction on inclines, which $M_1$ and $M_2$ cannot express. However, this modeling accuracy comes at the price of extremely high computation cost caused by Environmental Perception Bottlenecks (due to considering different terrain patches during every model query, rather than only computing based on the four wheel-terrain contact points in $M_1$ and $M_2$) and State Space Dimensionality (due to querying a full 6-DoF model, instead of efficient but inaccurate state space decomposition in $M_1$ and $M_2$). 
Like $M_2$, the trajectory rollout is inherently sequential: each step requires the previous prediction as part of its input state, preventing batch-parallel evaluation across timesteps and suffering from Sequential Time Dependencies.

\section{Experiments and Results}

\subsection{Experiment Setup}

We deploy our framework on custom wheeled platforms, utilizing mechanically capable 1/10th-scale, four-wheeled off-road vehicles. These vehicles feature all-wheel drive, independent suspensions, and differential locks, providing the necessary mechanical capability to traverse the vertically challenging terrain of a custom-built indoor testbed. To support our vision-based semantic mapping, the robots are equipped with a Microsoft Azure Kinect RGB-D camera. Onboard computation is handled by the NVIDIA Jetson Orin NX platform, which provides the required acceleration to run both the feature extraction and the sampling-based MPPI planner in real time.

\begin{figure*}[htbp]
    \centering
    \begin{subfigure}[b]{0.32\textwidth}
        \centering
        \includegraphics[width=\linewidth]{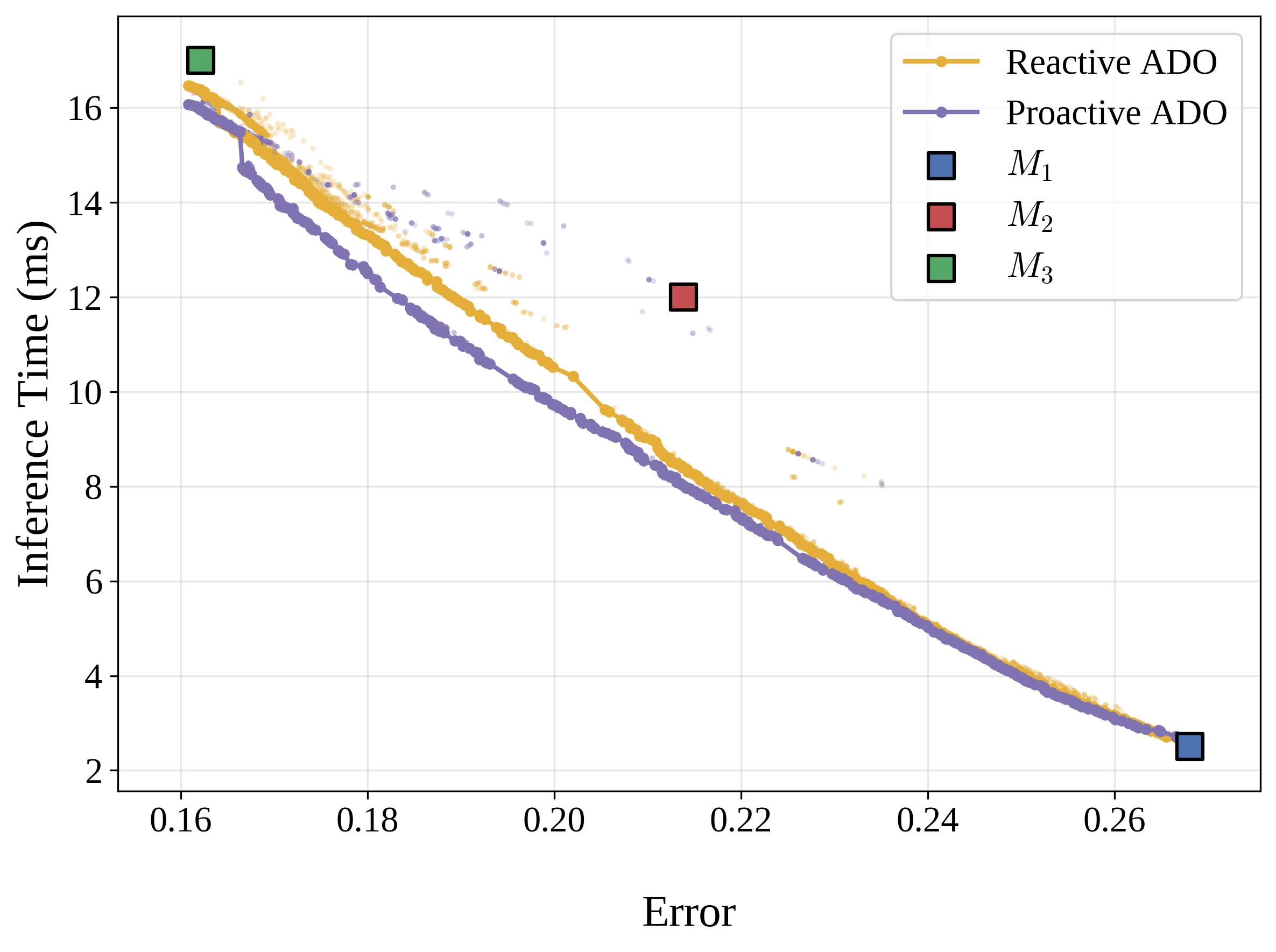}
        \caption{Pareto Front}
        \label{fig:pareto}
    \end{subfigure}
    \hfill
    \begin{subfigure}[b]{0.32\textwidth}
        \centering
        \includegraphics[width=\linewidth]{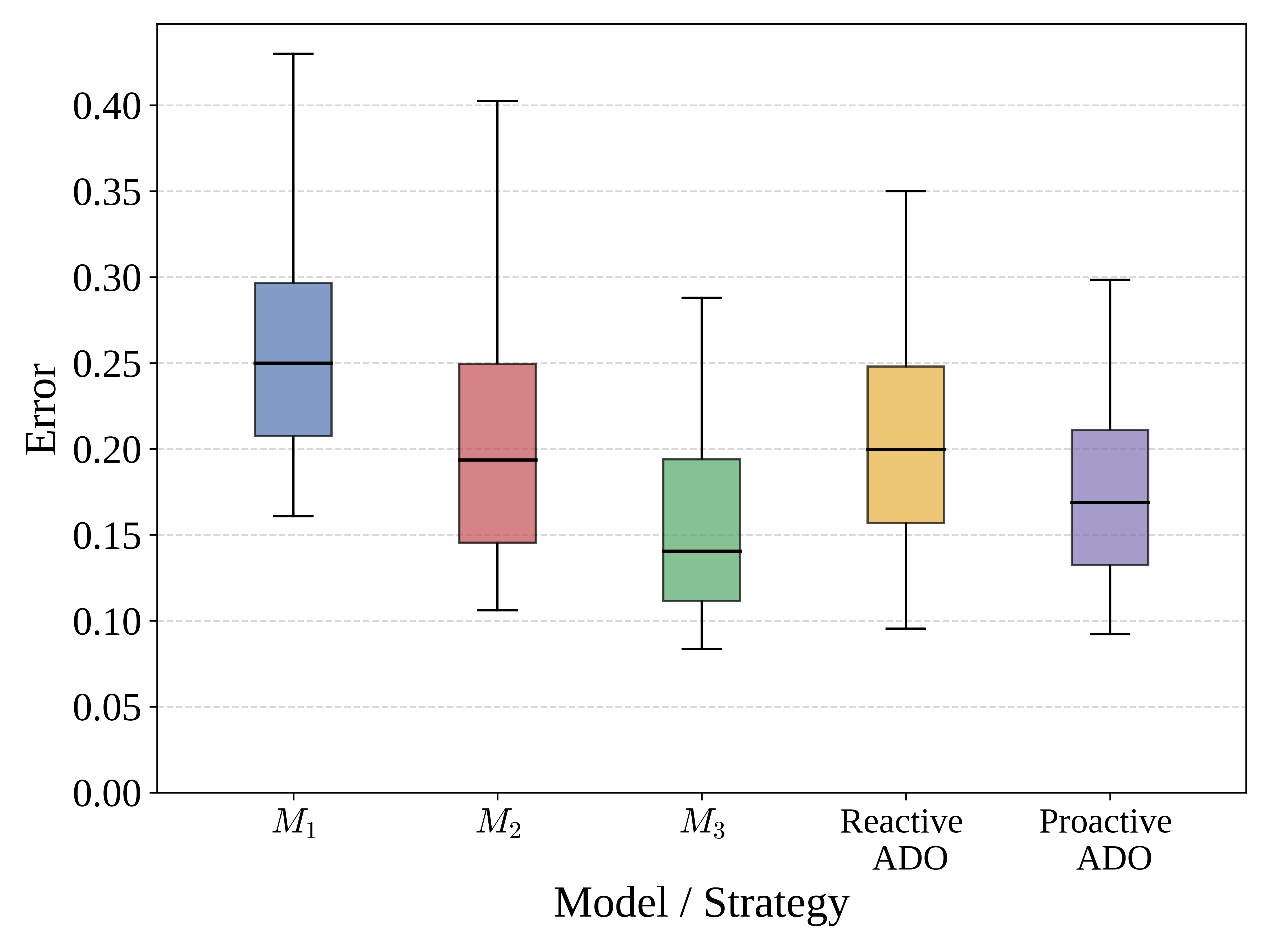}
        \caption{Total Error}
        \label{fig:error}
    \end{subfigure}
    \hfill
    \begin{subfigure}[b]{0.32\textwidth}
        \centering
        \includegraphics[width=\linewidth]{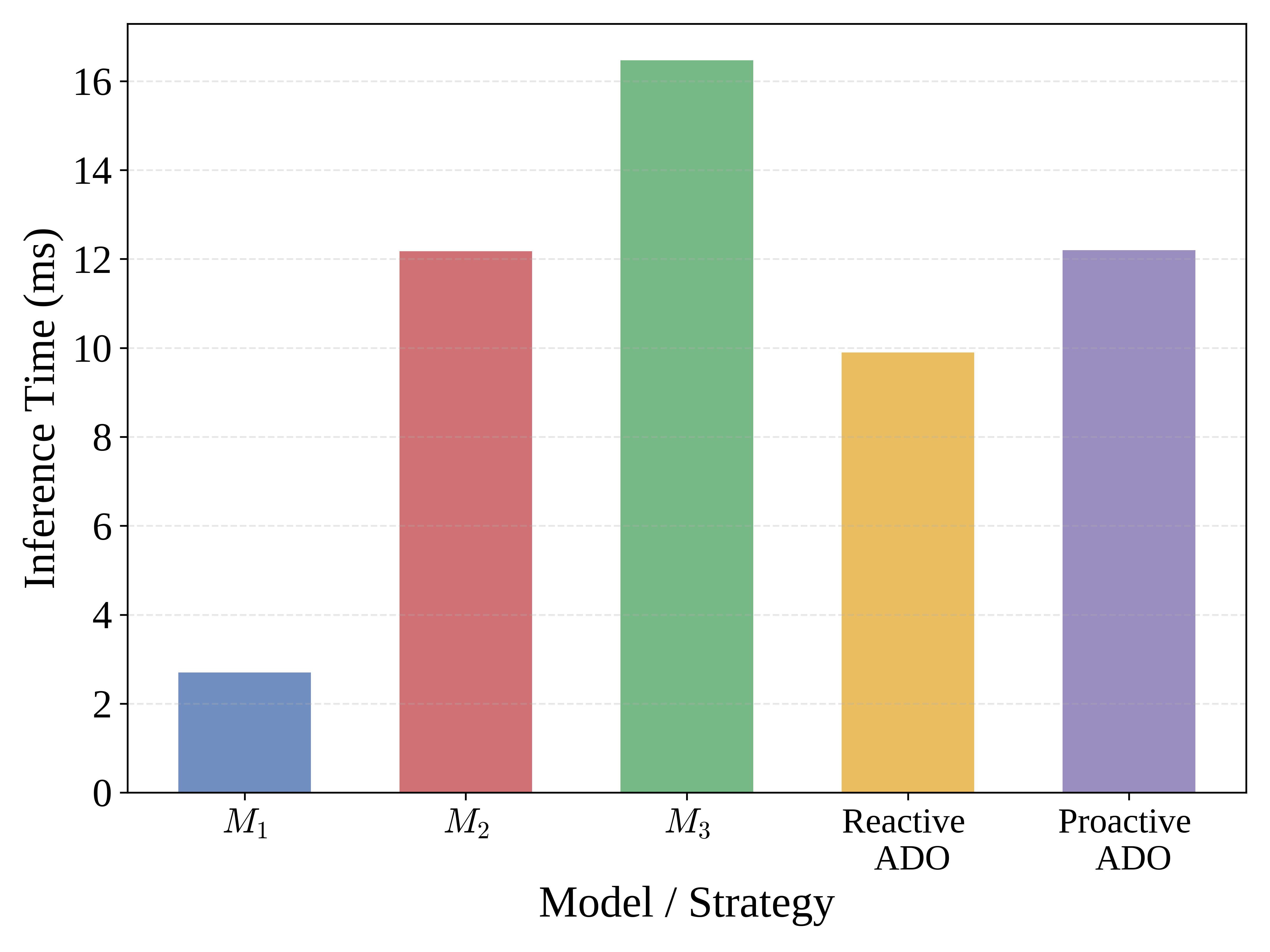}
        \caption{Inference Time}
        \label{fig:latency}
    \end{subfigure}
    \caption{Comparison of optimization results across different metrics.}
    \label{fig:combined_results}
\end{figure*}

\begin{figure}
    \centering
    \includegraphics[width=\linewidth]{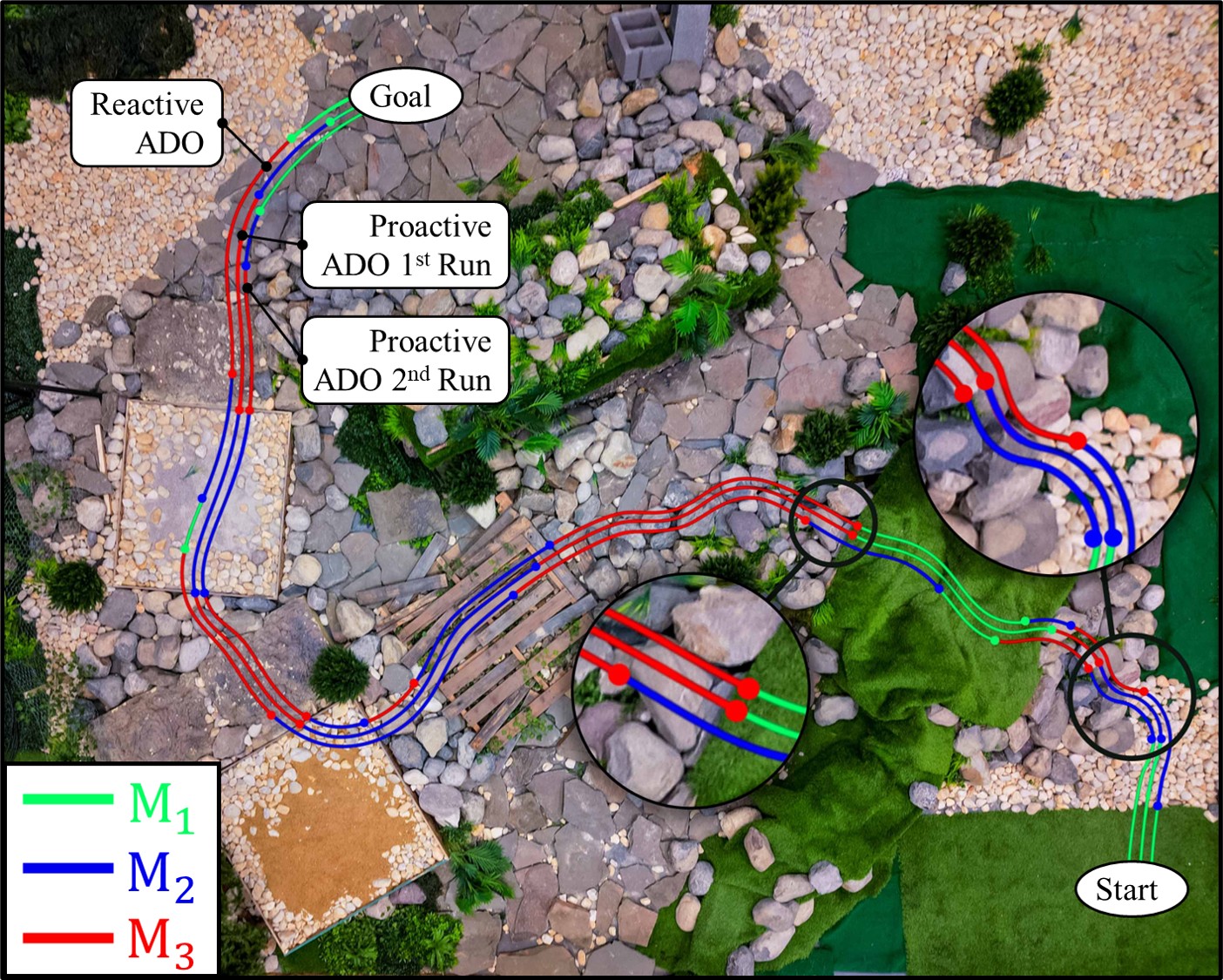}
    \caption{\our switching points for Reactive \our (left), Proactive \our without knowledge (middle), and Proactive \our with previous  knowledge (right). Proactive \our switches earlier before difficult terrain and further improves its switching behavior through online learning across runs.}
    \label{fig:arena}
\end{figure}

To evaluate \our, we utilize a highly configurable 8$\times$8 m indoor testbed (Fig.~\ref{fig:arena}) designed for off-road autonomy. The testbed presents a variety of vertically challenging conditions with elevation differences up to 0.7 m. It incorporates a continuous blend of distinct semantic terrain types, including deformable surfaces like sand and stone dust, as well as rigid obstacles like boulders, concrete, and varied grass. This rich semantic and geometric diversity is critical for validating the vision-based terrain identification and predictive semantic module. Furthermore, the testbed is enclosed by an eight-camera motion capture system that provides high-precision ground-truth localization.

\begin{table}[h]
\centering
\caption{Parameter search grid.}
\label{tab:param_grid}
\begin{tabular}{l c c c}
\toprule
Parameter & Min & Max & Step ($\delta$) \\
\midrule
$\alpha$ & 0.0 & 1.0 & 0.01 \\
$\beta$  & 0.0 & 3.0 & 0.1 \\
$\gamma$ & 0.0 & 1.0 & 0.01 \\
\bottomrule
\end{tabular}
\end{table}

\subsection{Parameter Estimation}

The ideal parameters ($\alpha, \beta, \gamma$) from Eqn.~\eqref{eq:score} for \our should favor simple models when the terrain is easy and only switch to complex models when accuracy gains justify the added compute. 
To avoid exhaustive real-world tuning in order to find these parameters, we leverage the counterfactual nature of the system.
Because a model's prediction error $e$ depends entirely on the sequence of past states and actions, it does not matter which model was actively controlling the robot at the time. By taking a recorded log of the robot driving over mixed terrain, we can replay the executed actions through every model in our library. This lets us calculate the prediction error and processing time each model would have produced given that identical history. Consequently, we can simulate how different parameter configurations would have altered the model-switching behavior. This allows us to rapidly perform an offline grid search over $(\alpha,\beta,\gamma)$ (Table~\ref{tab:param_grid}) to find the ideal balance between accuracy and efficiency.

Fig.~\ref{fig:combined_results}a plots these outcomes for the complete grid sweep and highlights the Pareto-optimal envelope (lower error vs.\ lower inference time).
As expected, configurations that heavily favor fast models collapse toward the low inference time endpoint (near $M_1$) but incur larger prediction error, while accuracy-dominant settings move toward the high-fidelity endpoint (near $M_3$) at substantially higher compute cost.
Crucially, the Pareto front exhibits a ``knee'' where the accuracy gains from \our outweigh the added latency. This is especially clear near $M_2$, where \our achieves substantially lower error for the same average inference time. Furthermore, Proactive \our outperforms Reactive \our by anticipating terrain changes and switching early, thereby avoiding the accumulated error that the reactive approach must build up before triggering a model change.

Having established the existence of a meaningful trade-off, we next evaluate a single operating point by choosing the hyperparameters that minimize the error as well as inference time on the traversal.
We then compare this tuned adaptive policy against deploying each model $M_1$, $M_2$, and $M_3$ in isolation.
Fig.~\ref{fig:combined_results}b shows that the tuned selector attains lower error by 30\% than $M_2$, whilst Fig.~\ref{fig:combined_results}c shows that it avoids the computational burden of always running the most expensive model.
Together, Fig.~\ref{fig:combined_results}b--c demonstrate the central claim of this section: with one tuned parameter set, the adaptive selector achieves a better accuracy-efficiency operating point than any fixed choice of $M_1$, $M_2$, or $M_3$, validating the value of \our rather than committing to a single model globally.

\subsection{Real-World Experiments}

\begin{table}[h]
\centering
\caption{Comparison of kinodynamic models and \our strategies.}
\resizebox{\linewidth}{!}{%
\begin{tabular}{lccccc}
\toprule
Metric & $M_1$ & $M_2$ & $M_3$ & Reactive \our & Proactive \our \\
\midrule
Traversal Time (s) $\downarrow$& 39.8 & \textbf{35.1} & 58.0 & 43.6 & 37.0 \\
Error $e$ $\downarrow$& 0.464 & 0.314 & \textbf{0.169} & 0.239 & 0.212 \\
Variance $\downarrow$ & 0.051 & 0.030 & \textbf{0.006} & 0.022 & 0.007 \\
Inference Time (ms) $\downarrow$& \textbf{2.70} & 12.5 & 17.4 & 12.6 & 12.8 \\
Success rate  $\uparrow$ & 5/10 & 8/10 & \textbf{10/10} & \textbf{10/10} & \textbf{10/10} \\
\bottomrule
\end{tabular}%
}
\label{tab:real_exp}
\end{table}

\subsubsection{Physical Deployment and Benchmarking}
The offline replay study in Sec.~VI-B identifies a single operating point that balances prediction fidelity against inference cost.
We validate that this operating point transfers to real deployment by running the complete system on the physical platform using the same hyperparameters and benchmarking it against fixed-model baselines.

We execute the full MPPI stack with Reactive and Proactive \our enabled and compare against three static deployments ($M_1$, $M_2$, and $M_3$ alone).
Rather than degrading the MPPI optimization quality for computationally expensive models by reducing the sample count or horizon length to claw back compute time, we apply a kinematic penalty: the maximum linear and angular velocities for the static baselines are scaled inversely to their average computational inference time.

Each method is evaluated over multiple repeated runs on the same mixed-terrain course, and Table~\ref{tab:real_exp} reports success rate, traversal time, tracking error statistics, and average inference time.
The reported average inference time for the \our strategies represents the total effective latency per control cycle, including the overhead of the visual-semantic pipeline, the asynchronous learning loop updates, and the counterfactual evaluation of the inactive models.

Table~\ref{tab:real_exp} demonstrates that \our achieves a favorable accuracy-efficiency trade-off in the real system.
The fastest baseline ($M_1$) attains low runtime but incurs substantially higher tracking error, while the highest-fidelity baseline ($M_3$) reduces error at the cost of increased latency and longer traversal time.
In contrast, \our attains traversal times comparable to the faster baselines while significantly reducing tracking error, closing much of the gap toward $M_3$ without paying its computational cost uniformly.
This confirms that selectively allocating model complexity yields improved end-to-end performance relative to any single fixed model.

Table~\ref{tab:real_exp} further shows that incorporating semantic context in proactive \our improves robustness in deployment.
The semantic-aware variant reduces both the mean tracking error and its variance, indicating fewer high-error excursions near transitions, while adding only marginal computational overhead.
Overall, these real-world results corroborate the offline analysis and validate \our as a practical mechanism for achieving low error under real-time constraints.

\subsubsection{Online Adaptation on Complex Terrain}

To further evaluate the online learning capabilities of the framework, we conduct a sequential traversal experiment targeting the most challenging terrain in the testbed (Fig. \ref{fig:arena}). This specific course section features extreme elevation differences up to 0.7 m and a continuous blend of distinct semantic terrain types, switching from rocks to grass and back. By focusing on this difficult segment, we can clearly observe the performance difference of reactive switching, zero-shot predictive switching, and experienced predictive switching.

We execute three distinct runs over the identical path. First, we deploy the Reactive \our configuration, which adjusts the model preference only after the robot has already traversed new terrain and accumulated substantial prediction error. Next, we run Proactive \our without prior experience on this specific path, relying on the visual-semantic pipeline being trained from scratch to anticipate environmental changes. Finally, we perform a second run of Proactive \our over the same route. In this final traversal, the asynchronous learning loop has already updated the historical mean error $E_{i,k}$ and variance $V_{i,k}$ based on the visual context recorded during the previous run.

\begin{table}[htbp]
\centering
\caption{Comparison of online adaptation strategies on complex terrain.}
\label{tab:online_adaptation}
\resizebox{\linewidth}{!}{%
\begin{tabular}{l c c c}
\toprule
Strategy & Reactive \our & Proactive \our & Proactive \our \\
\midrule
Prior Knowledge & None & None & Full \\
Traversal Time (s) $\downarrow$ & 44.3 & 39.1 & \textbf{35.2} \\
Error $e$ $\downarrow$ & 0.247 & 0.229 & \textbf{0.211} \\
Variance $\downarrow$ & 0.021 & 0.016 & \textbf{0.008} \\
Inference Time (ms) $\downarrow$ & \textbf{11.7} & 12.2 & 12.4 \\
\bottomrule
\end{tabular}%
}
\end{table}
The results in Table~\ref{tab:online_adaptation} demonstrate a clear progression in navigation performance. Introducing the semantic layer during the first proactive run reduces tracking error compared to the reactive baseline by anticipating future terrain changes and proactively switching models before the tracking error grows. However, the most significant improvement is observed during the second proactive traversal. By continuously refining terrain-conditioned performance estimates using residual errors from online counterfactual rollouts, the system leverages its acquired knowledge to achieve highly accurate predictive gating. This confirms that the framework effectively learns on the fly to maximize closed-loop performance under real-time constraints.

\section{Conclusion}

We present \our, an adaptive model-orchestration framework that selects among a library of dynamics models with different accuracy-efficiency profiles to maximize closed-loop performance under real-time constraints.
Rather than committing to a single model globally, \our treats model choice as a control-relevant decision and leverages terrain context (including semantics) to allocate computation where it matters most.

Across offline counterfactual replay and repeated real-world trials, \our consistently achieves a superior accuracy--efficiency operating point relative to fixed-model baselines.
The selector preserves throughput on benign terrain by favoring inexpensive models, and escalate to higher-fidelity models on difficult segments to reduce tracking error.
Proactively incorporating semantic context further improves robustness by reducing error variance and mitigating high-error excursions near terrain transitions, while introducing only marginal computational overhead.

While ADO optimizes the accuracy-efficiency trade-off, it currently relies on soft clustering for terrain categorization, which requires predefined classes and lacks robustness in out-of-distribution environments. Furthermore, the framework has only been validated on a single 1/10th-scale wheeled platform with a library restricted to three dynamics models.
Future research will focus on transitioning from discrete soft clustering toward continuous latent embeddings for flexible model selection in novel environments. Additionally, ADO needs to be evaluated across diverse platforms, such as legged robots, to ensure cross-system portability. Finally, coupling model selection with adaptive MPC parameters, such as sample density and horizon length, for holistic compute allocation is another promising future research direction.

\section*{ACKNOWLEDGMENT}
This work has taken place in the RobotiXX Laboratory at George Mason University. RobotiXX research is supported by National Science Foundation (NSF, 2350352), Army Research Office (ARO, W911NF2320004, W911NF2520011), Google DeepMind (GDM), Clearpath Robotics, FrodoBots Lab, Raytheon Technologies (RTX), Tangenta, Mason Innovation Exchange (MIX), and Walmart.

F. Cancelliere and G. Muscato acknowledge financial support from PNRR MUR project PE0000013-FAIR.

\bibliographystyle{IEEEtran}
\bibliography{IEEEabrv,references}

\end{document}